\documentclass[10pt,a4paper,onecolumn]{article}
\usepackage[a4paper, total={6in, 8in}]{geometry}
\usepackage{setspace,url}
\usepackage{lineno,hyperref,multirow,cite}
\usepackage{authblk}
\usepackage{epsfig,amssymb,amsmath}
\usepackage[utf8]{inputenc}
\usepackage{amsmath,graphicx}
\usepackage{amssymb,amsmath,epsfig,url,mathrsfs} \usepackage{algorithm,algorithmic}
\usepackage[usenames, dvipsnames]{color}
\usepackage[framemethod=TikZ]{mdframed}
\usepackage{xcolor}
\usepackage{colortbl}
\usepackage[colorinlistoftodos]{todonotes}
\usepackage{multirow}
\usepackage{multicol}
\usepackage{enumitem}

\usepackage{numprint}
\npthousandsep{,}

\usepackage{graphicx}
\usepackage{subfig}
\usepackage{array, makecell}
\usepackage{verbatim}

\usepackage[utf8]{inputenc}
\usepackage[T1]{fontenc}
\usepackage[ngerman,english]{babel}
\usepackage{tabularx}  % for 'tabularx' environment and 'X' column type
\usepackage{ragged2e}  % for '\RaggedRight' macro (allows hyphenation)
\newcolumntype{Y}{>{\RaggedRight\arraybackslash}X} 
\usepackage{booktabs}  % for \toprule, \midrule, and \bottomrule macros 
\usepackage{amsthm}

\makeatletter
\newcommand{\mathleft}{\@fleqntrue\@mathmargin0pt}
\newcommand{\mathcenter}{\@fleqnfalse}
\makeatother
 
\theoremstyle{definition}

\hypersetup{
   unicode=false,          % non-Latin characters in Acrobat’s bookmarks
   pdftoolbar=true,        % show Acrobat’s toolbar?
   pdfmenubar=true,        % show Acrobat’s menu?
   pdffitwindow=false,     % window fit to page when opened
   pdfstartview={FitH},    % fits the width of the page to the window
   pdftitle={VoicePrivacy 2020 Challenge Evaluation Plan},    % title
   pdfauthor={VoicePrivacy},     % author
   pdfsubject={VoicePrivacy},   % subject of the document
   pdfcreator={},   % creator of the document
   pdfproducer={}, % producer of the document
   pdfkeywords={VoicePrivacy, anonymisation}, % list of keywords
   pdfnewwindow=true,      % links in new window
   colorlinks=true,       % false: boxed links; true: colored links
   linkcolor=blue,          % color of internal links
   citecolor=blue,        % color of links to bibliography
   filecolor=magenta,      % color of file links
   urlcolor=blue           % color of external links
}

\usepackage{color, colortbl}
\definecolor{LightCyan}{rgb}{0.88,1,1}

\mdfdefinestyle{MyFrame}{%
    linecolor=black,
    outerlinewidth=.1pt,
    roundcorner=1pt,
    innertopmargin=\baselineskip,
    innerbottommargin=\baselineskip,
    innerrightmargin=5pt,
    innerleftmargin=-10pt,
    backgroundcolor=gray!10!white}

\title{The VoicePrivacy 2020 Challenge\\ Evaluation Plan\\[1em]\large{}Version 1.4}

\author[1]{Natalia Tomashenko}
\author[2]{Brij Mohan Lal Srivastava}
\author[3]{Xin Wang}
\author[4]{Emmanuel Vincent}
\author[5]{Andreas Nautsch}
\author[3,6]{Junichi Yamagishi}
\author[5]{Nicholas Evans}
\author[5]{Jose Patino}
\author[1]{Jean-François Bonastre}
\author[1]{Paul-Gauthier Noé}
\author[5]{Massimiliano Todisco}
\affil[1]{Laboratoire Informatique d’Avignon (LIA), Avignon Université, France}
\affil[2]{Inria, France}
\affil[3]{National Institute of Informatics, Tokyo, Japan}
\affil[4]{Université de Lorraine, CNRS, Inria, LORIA, France}
\affil[5]{Audio Security and Privacy Group, EURECOM, France}
\affil[6]{University of Edinburgh, UK}

\date{\url{https://voiceprivacychallenge.org}}

\begin{document}

\maketitle

\begin{abstract}
    The VoicePrivacy Challenge aims to promote the development of privacy preservation tools for speech technology by gathering a new community to define the tasks of interest and the evaluation methodology, and benchmarking solutions through a series of challenges. In this document, we formulate the voice anonymization task selected for the VoicePrivacy 2020 Challenge and describe the datasets used for system development and evaluation. We also present the attack models and the associated objective and subjective evaluation metrics. We introduce two anonymization baselines and report objective evaluation results.
\end{abstract}

\section{Context}

Recent years have seen mounting calls for the preservation of privacy when treating or storing personal data. This is not least the result of recent European privacy legislation, e.g., the general data protection regulation (GDPR). While there is no legal definition of privacy~\cite{nautsch2019gdpr}, speech data is likely to fall within the scope of privacy regulation.  This is because speech data contains much more than just the spoken words. Speech encapsulates a wealth of personal, personal data, e.g., age and gender, health and emotional state, racial or ethnic origin, geographical background, social identity, and socio-economic status~\cite{Nautsch-PreservingPrivacySpeech-CSL-2019}.  Speaker recognition systems can also reveal the speaker's identity. Since machines can now decipher spoken language with ever-impressive accuracy, there is no reason why political orientations, religious and philosophical beliefs could not also be derived from speech data.

Given that speech data can qualify as personal data, it is thus of no surprise that efforts to develop privacy preservation solutions for speech technology are starting to emerge.  The study of privacy is also gaining traction; the Security and Privacy in Speech Communication (SPSC)\footnote{\url{https://www.spsc-sig.org}} Special Interest Group (SIG) of the International Speech Communication Association (ISCA) has recently been formed to champion research in the domain.  

There are arguably two general solutions to preserve privacy in speech data: cryptography, and anonymization, also referred to as de-identification.  Cryptographic solutions can render speech data unreadable and can only be read in plain text with access to a private key. Some techniques, e.g.\ homomorphic encryption, even support computation upon data in the encrypted domain.  These methods are typically specific to the given application, are challenging to integrate within existing systems and almost always result in significant increases to computational complexity and/or communication overheads.  As such, cryptographic solutions can be relatively inflexible and cumbersome.  Even then, they can only be implemented with specialist cryptography expertise. 

Anonymization is different.  Instead of conserving and protecting sensitive data  through encryption, anonymization techniques, e.g.~\cite{fang2019speaker}, have the goal of suppressing personally identifiable information within a speech signal, leaving all other aspects of the speech signal intact. Anonymization techniques can be highly flexible and can be integrated easily within existing systems. They usually have low computational complexity and rarely require any additional expertise outside of the speech domain.  Despite the appeal of anonymization techniques and the urgency to address privacy concerns, there are currently only few solutions.  In addition, a formal definition of anonymization is missing, the level of anonymization offered by some solutions is somewhat unclear, and there are no common datasets, protocols and metrics to support further work.

The VoicePrivacy initiative is spearheading the effort to develop privacy preservation solutions for speech technology. It aims to gather a new community to define the task and metrics and to benchmark initial solutions using the first common datasets and protocols.  In following best practice, VoicePrivacy takes the form of a competitive challenge.
Challenge participants are required to process a dataset of speech signals in order to 
anonymize them, while protecting the linguistic content and speech naturalness.
The challenge will run from early 2020 and conclude with a special session/event held in conjunction with Interspeech 2020, at which challenge results will be made publicly available.  This document describes plans for the challenge, the dataset, protocol and the set of metrics that will be used for assessment, in addition to evaluation rules and guidelines for registration and submission.

\section{Challenge objectives}
The grand objective of the VoicePrivacy challenge is to foster progress in the development of anonymization techniques for speech data.  The specific technical goals are summarised as follows.  They are to:

\begin{itemize}
\item facilitate the development of novel techniques which suppress speaker-discriminative information within speech signals;% that is used by automatic systems to recognise speakers;
\item promote techniques which provide effective anonymization while protecting intelligibility and naturalness;
\item provide a level playing field to facilitate the comparison of different anonymization solutions using a common dataset and protocol;
\item investigate metrics for the evaluation and meaningful comparison of different anonymization solutions.
\end{itemize}

VoicePrivacy participants will be provided with a common datasets, protocols and a suite of software packages that will enable them to gauge anonymization performance.  While it is likely that this will evolve in future challenges, no formal ranking of systems will be made for this first edition.  This is because recommendations for the assessment of anonymization solutions form one of the technical goals.

\section{Task}
\label{sec:task}
A privacy preservation task is typically formulated as a game involving one or more \emph{users} who publish some data and an \emph{attacker} (also called an adversary) who gains legal or illegal access to this data or to derived data and attempts to infer personal information about the users. To protect their privacy, the users publish data that contain as little personal information as possible while allowing one or more desired goals to be fulfilled. To infer personal information about the users, the attacker may use additional knowledge.

Considering speech data as an example, the definition of the privacy preservation task depends on the following scenario specifications~\cite{srivastava2019evaluating}:
\begin{itemize}
\item the nature of the data (e.g., speech waveform, speech features, text, etc.)
\item the pieces of information considered as personal (e.g., speaker identity, speaker traits, spoken contents, etc.)
\item the desired goal(s) (e.g., human dialogue, multi-party human conversation, automatic speech recognition (ASR) training, emotion recognition, etc.)
\item the data accessed by the attacker (e.g., a single utterance, multiple utterances, features computed from these utterances, an ASR model trained on these utterances, etc.)
\item the additional knowledge available to the attacker (e.g., utterances previously released by the user, the measures adopted by users to protect their privacy, etc.)
\end{itemize}

Different scenario specifications lead to different privacy preservation methods from the user's point of view and different attack methods from the attacker's point of view.

\subsection{Scenario}
In VoicePrivacy 2020, we consider the following specific scenario, where the terms ``user'' and ``speaker'' are used interchangeably:
\begin{itemize}
\item Speakers want to hide their identity while allowing any desired goal to be potentially achieved.
\item The attacker has access to a single utterance and wants to identify the corresponding speaker. 
\end{itemize}

\subsection{Anonymization task}
\label{subsec:user_goals}
In order to hide his/her identity, each speaker passes his/her utterances through an anonymization system before publication. The resulting anonymized utterances are called \emph{trial} utterances. They sound as if they were uttered by another speaker, which we call \emph{pseudo-speaker} which may be an artificial voice not corresponding to any real speaker.

The task of challenge participants is to develop this anonymization system. In order to allow any desired goal to be potentially achieved, this system should satisfy the following requirements:
\begin{itemize}
    \item output a speech waveform,
    \item hide speaker identity as much as possible,
    \item distort other speech characteristics as little as possible,
    \item ensure that all trial utterances from a given speaker appear to be uttered by the same pseudo-speaker, while trial utterances from different speakers appear to be uttered by different pseudo-speakers.
\end{itemize}

The third requirement will be assessed via a range of \emph{utility} metrics (also called usability metrics). Specifically, ASR performance using a model trained on clean (non-anonymized) data and subjective speech intelligibility and naturalness will be measured during the challenge (see Section \ref{sec:perf_objective} and \ref{sec:perf_subjective}), and additional desired goals including ASR training will be assessed in a post-evaluation stage (see Section \ref{sec:posteval}).

The fourth requirement is motivated by the fact that, in a multi-party human conversation, each speaker cannot change his/her anonymized voice over time and the anonymized voices of all speakers must be distinguishable from each other. This will be assessed via a new, specifically designed metric in the post-evaluation stage.

\subsection{Attack model}
\label{subsec:attack_model}
We assume that the attacker has access to various amounts of data:
\begin{itemize}
\item one or more anonymized trial utterances,
\item possibly, several additional utterances for each speaker, which may or may not have been anonymized and are called \emph{enrollment} utterances.
\end{itemize}
The attacker does not have access to the anonymization system applied by the user though.

The level of protection of personal information will be assessed via a range of \emph{privacy} metrics, including objective speaker verifiability metrics and subjective speaker verifiability and linkability metrics. These metrics assume different attack models:
\begin{itemize}
    \item The objective speaker verifiability metrics assume that the attacker has access to a single anonymized trial utterance and several enrollment utterances. During the challenge, two sets of metrics will be computed, corresponding to the two situations when the enrollment utterances are clean or they have been anonymized (see Section \ref{sec:perf_objective}). In the latter case, we assume that the utterances have been anonymized in the same way as the trial data using the same anonymization system, i.e., all enrollment utterances from a given speaker are converted into the same pseudo-speaker, and enrollment utterances from different speakers are converted into different pseudo-speakers. We also assume that the pseudo-speaker corresponding to a given speaker in the enrollment set is different from the pseudo-speaker corresponding to that same speaker in the trial set. In the post-evaluation stage, we will consider alternative anonymization procedures for the enrollment data, corresponding to stronger attack models (see Section \ref{sec:posteval}).
    \item The subjective speaker verifiability metric assumes that the attacker has access to a single anonymized trial utterance and a single clean enrollment utterance (see Section \ref{sec:perf_subjective}).
    \item The subjective speaker linkability metric assumes that the attacker has access to  several anonymized trial utterances 
    (see Section \ref{sec:perf_subjective}).
\end{itemize}

\section{Data}

Several publicly available corpora are used for training, development and evaluation of speaker anonymization systems.  They are comprised of subsets from the following corpora:
\begin{itemize}
    \item \textit{\textbf{LibriSpeech}}\footnote{Librispeech: \url{http://www.openslr.org/12}}~\cite{panayotov2015librispeech} is a corpus of read English speech derived from audiobooks and designed for
ASR research. It contains about \numprint{1000} hours of speech sampled at 16 kHz.

    \item \textit{\textbf{LibriTTS}}\footnote{LibriTTS: \url{http://www.openslr.org/60/}}~\cite{zen2019libritts} is a corpus of English speech derived from the original LibriSpeech corpus and designed
for text-to-speech (TTS). It contains approximately 585~hours of read English speech sampled at 24 kHz.

    \item {\textbf{\textit{VCTK}}}\footnote{VCTK, release version 0.92: \url{https://datashare.is.ed.ac.uk/handle/10283/3443}}~\cite{yamagishi2019cstr} is a corpus of read speech of 109 native speakers of English with various accents. It was originally aimed for  TTS synthesis systems and contains about 44 hours of speech sampled at 48~kHz.

    \item \textit{\textbf{VoxCeleb-1,2}}\footnote{VoxCeleb: \url{http://www.openslr.org/60/}}~\cite{nagrani2017voxceleb,chung2018voxceleb2} is an audiovisual corpus extracted from videos uploaded to YouTube and designed for speaker verification research. It contains
    about \numprint{2770} hours of speech (16 kHz) from about \numprint{7360} speakers, covering a wide range of accents and languages.

\end{itemize}

The detailed description of datasets provided for training, development and evaluation is given below.

\subsection{Training data}\label{sec:train-data}

For training an anonymization system the following corpora can be used:
\textit{VoxCeleb-1,2}, \textit{LibriSpeech-train-clean-100}, \textit{LibriSpeech-train-other-500},
\textit{LibriTTS-train-clean-100} and 
\textit{LibriTTS-train-other-500}. 
Summary statistics for these data are given in Table~\ref{tab:data-train}.

\begin{table}[htbp]
  \caption{Statistics of the \textbf{training} datasets.}\label{tab:data-train}
  \renewcommand{\arraystretch}{1.1}
  \centering
  \begin{tabular}{|l|r|r|r|r|r|}
\Xhline{0.6pt}
   \multirow{2}{*}{\textbf{Subset}} &   \multirow{2}{*}{\textbf{Size,h}} & \multicolumn{3}{c|}{\textbf{Number of Speakers}} &  \multirow{2}{*}{\textbf{Number of Utterances}} \\ \cline{3-5}
  &  & \textbf{Female} & \textbf{Male} & \textbf{Total} & \\ \hline \hline
  VoxCeleb-1,2  & \numprint{2794} & \numprint{2912} & \numprint{4451} & \numprint{7363} &  \numprint{1281762} \\ \hline
  LibriSpeech: train-clean-100 & 	100	& 125 &	126 &	251	& \numprint{28539}	\\ \hline
  LibriSpeech: train-other-500 & 	497	& 564 &	602 &	\numprint{1166}	& \numprint{148688}	\\ \hline
  LibriTTS: train-clean-100 & 	54	& 123 &	124 &	247	& \numprint{33236}	\\ \hline
  LibriTTS: train-other-500 & 	310	& 560 &	600 &	\numprint{1160} & \numprint{205044}		\\ \Xhline{0.6pt}
  \end{tabular}
\end{table}

\subsection{Development set}\label{sec:data-dev}

Subsets from two different corpora are provided as development sets for anonymization systems.
The first one is a subset of the \textit{LibriSpeech-dev-clean} dataset. 
The second one (denoted as \textit{VCTK-dev}) is obtained from the VCTK corpus.
As explained in Section \ref{sec:task}, we split these datasets into a trial subset, which is to be anonymized, and an enrollment subset, which is to be anonymized with different pseudo-speakers in order to compute objective speaker verification metrics. The anonymized enrollment data shall be stored in a different folder than the clean enrollment data, since both clean and anonymized enrollment data will be used to computed two different sets of metrics.

The statistics of these subsets are summarized in Table~\ref{tab:data-dev-asv}. For the \textit{LibriSpeech-dev-clean} dataset, the speakers in the enrollment set are a subset of those in the trial set. For the \textit{VCTK-dev} dataset, we created two subsets of trial utterances, denoted as \textit{common part} and \textit{different part}.  
Both include trials from the same set of speakers, but from disjoint subsets of utterances.
The \textit{common part} of the trials is composed of utterances $\#1-24$ in the VCTK corpus that are identical for all speakers:  the \textsl{elicitation paragraph}\footnote{ Elicitation paragraph: \url{http://accent.gmu.edu/pdfs/elicitation.pdf}} (utterances $\#1-5$) and   \textsl{rainbow passage}\footnote{Rainbow passage: \url{https://www.dialectsarchive.com/the-rainbow-passage}} (utterances $\#6-24$).
This part is meant for subjective evaluation of speaker verifiability/linkability in a text-dependent manner.
The enrollment subset and the \textit{different part} of the trials are composed of distinct utterances for all speakers (utterances with indexes $\geq25$).

\begin{table}[htbp]
  \caption{Statistics of the \textbf{development} datasets. }\label{tab:data-dev-asv}
  \renewcommand{\arraystretch}{1.1}
  \centering
  \begin{tabular}{|l|l|r|r|r|}
\Xhline{0.6pt}
 \textbf{Subset} & &  \textbf{Female} & \textbf{Male} & \textbf{Total}  \\ \hline \hline
\multirow{4}{*}{LibriSpeech: dev-clean}  &  Speakers in enrollment & 15 &	14	& 29 \\ \cline{2-5}
 &  Speakers in trials & 20 &	20	& 40 \\ \cline{2-5}
& Enrollment utterances & 167 &	176	& 343 \\  \cline{2-5}
& Trial utterances & \numprint{1018} &	960 	& \numprint{1978} \\ \hline \hline
\multirow{4}{*}{VCTK-dev}  &  Speakers (same in enrollment and trials) & 15 &	15	& 30 \\ \cline{2-5}
& Enrollment utterances & 300 &	300	& 600 \\  \cline{2-5}
& Trial utterances (common part) & 344 & 351	& 695\\  \cline{2-5}
& Trial utterances (different part) & \numprint{5422} &	\numprint{5255}	& \numprint{10677} \\  \Xhline{0.6pt}
  \end{tabular}
\end{table}

\subsection{Evaluation data}

Similarly to the development data, subsets from two different corpora are used for evaluation.
The first one is a subset of the \textit{LibriSpeech-test-clean} dataset. 
The second one (denoted as \textit{VCTK-test}) is obtained from the VCTK corpus. 
We split those datasets into enrollment and trial subsets as summarized in Table~\ref{tab:data-test-asv}.
For the VCTK dataset, we created two subsets of trial utterances, denoted as \textit{common part} and \textit{different part}, in a similar manner as for the development set.

%\end{table}

\begin{table}[htbp]
  \caption{Statistics of the \textbf{evaluation} datasets.}\label{tab:data-test-asv}
   \renewcommand{\arraystretch}{1.1}
  \centering
  \begin{tabular}{|l|l|r|r|r|}
\Xhline{0.6pt}
 \textbf{Subset} &  &  \textbf{Female} & \textbf{Male} & \textbf{Total}  \\ \hline \hline
\multirow{4}{*}{LibriSpeech: test-clean}  &  Speakers in enrollment & 16 &	13	& 29 \\ \cline{2-5}
&  Speakers in trials & 20 &	20	& 40  \\ \cline{2-5}
& Enrollment utterances & 254 &	184	& 438 \\  \cline{2-5}
& Trial utterances & 734 &	762	& 1496 \\  \hline \hline
\multirow{4}{*}{VCTK-test}  & 
Speakers (same in enrollment and trials) & 15 &	15	& 30 \\ \cline{2-5}
& Enrollment utterances & 300 & 300 & 600 \\  \cline{2-5}
& Trial utterances (common part) & 346 & 354 & 700 \\ \cline{2-5}
%& Trial utterances (common part) & 346 & 354 & 700‬ \\\cline{2-5}
% & Trial utterances (different part) & \numprint{5328} & \numprint{5420} & \numprint{10748}‬ \\ \Xhline{0.6pt}
& Trial utterances (different part) &\numprint{5328} & \numprint{5420} & \numprint{10748}\\ \Xhline{0.6pt}
\end{tabular}
\end{table}

\section{Utility and privacy metrics}
\label{sec:perf}

Following the attack model assumed in Section \ref{subsec:attack_model}, we consider objective and subjective privacy metrics to assess speaker re-identification and linkability. We also propose objective and subjective utility metrics in order to assess the fulfillment of the user goals specified in Section \ref{subsec:user_goals}. Specifically, we consider ASR performance using a model trained on clean data and subjective speech intelligibility and naturalness.

Objective metrics will be computed by the participants themselves using the provided evaluation scripts and evaluation data, while subjective metrics will be computed the organizers using data provided by the participants.

\subsection{Objective metrics}
\label{sec:perf_objective}

For objective evaluation of anonymization performance, two systems will be trained to assess the following characteristics: 
(1) speaker verifiability and  (2)  ability of the anonymization system to  
  preserve linguistic information in the anonymized speech.
The first system denoted as \textrm{$ASV_\text{eval}$} is an automatic speaker verification (ASV) system, which produces log-likelihood ratio (LLR) scores.
The second system denoted as \textrm{$ASR_\text{eval}$} is an automatic speech recognition (ASR) system which outputs a word error rate (WER) metric.

Both \textrm{$ASR_\text{eval}$} and \textrm{$ASV_\text{eval}$} are trained on the \textit{LibriSpeech-train-clean-360} dataset using the Kaldi speech recognition toolkit~\cite{povey2011kaldi}. The statistics of this dataset are given in Table~\ref{tab:train-eval-metrics}.
Training of \textrm{$ASR_\text{eval}$} and \textrm{$ASV_\text{eval}$} and evaluation will be done with the provided  recipes\footnote{Evaluation scripts: \url{https://github.com/Voice-Privacy-Challenge/Voice-Privacy-Challenge-2020}}.

\begin{table}[htbp]
  \caption{Statistics of the training dataset for the \textrm{$ASV_\text{eval}$} and \textrm{$ASR_\text{eval}$ evaluation systems.}}\label{tab:train-eval-metrics}
  \renewcommand{\arraystretch}{1.1}
  \centering
  \begin{tabular}{|l|r|r|r|r|r|}
\Xhline{0.6pt}
   \multirow{2}{*}{\textbf{Subset}} &   \multirow{2}{*}{\textbf{Size,h}} & \multicolumn{3}{c|}{\textbf{Number of Speakers}} &  \multirow{2}{*}{\textbf{Number of Utterances}} \\ \cline{3-5}
  &  & \textbf{Female} & \textbf{Male} & \textbf{Total} & \\ \hline \hline
  LibriSpeech: train-clean-360 & 363.6 & 439 & 482	 &	921	& \numprint{104014}	\\ \Xhline{0.6pt}
  \end{tabular}
\end{table}

\subsubsection{Word error rate (WER)}\label{sec:wer}

\begin{figure}[tp]
\centering\includegraphics[width=82mm]{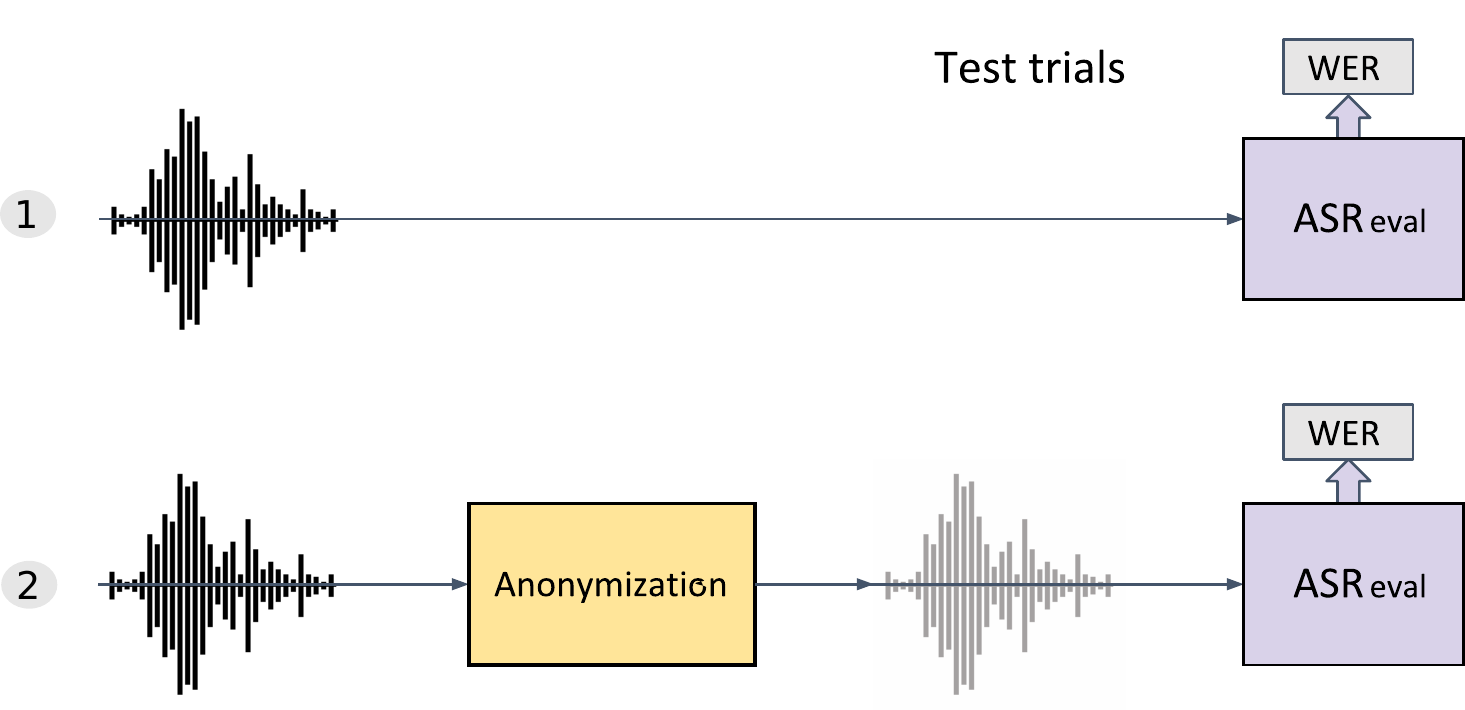}
\caption{ASR evaluation for (1) clean speech data and (2)  anonymized speech data. WER is estimated on the \textsl{trial} utterances of the development and evaluation datasets.}
\label{fig:asr-eval}
\end{figure}

ASR performance will be assessed using \textrm{$ASR_\text{eval}$} which is based on the state-of-the-art Kaldi recipe for LibriSpeech involving a TDNN-F acoustic model and a trigram language model.
The recipe has been adapted to run on the \textit{LibriSpeech-train-clean-360} dataset rather than the full LibriSpeech training corpus in order to speed up the evaluation process.

As shown in Figure~\ref{fig:asr-eval}, the anonymized development and evaluation data will be decoded using the provided pretrained system \textrm{$ASR_\text{eval}$} and the word error rate (WER) will be calculated: 
\begin{equation*}
\text{WER}=\frac{N_\text{sub}+N_\text{del}+N_\text{ins}}{N_\text{ref}}
\end{equation*}
with $N_\text{sub}$, $N_\text{del}$, and $N_\text{ins}$, the number of substitution, deletion, and insertion errors, respectively, and $N_\text{ref}$ the number of words in the reference. This will be compared to the WER achieved on clean data. The WER is calculated only on the trial part of the development and evaluation datasets.

 \begin{figure}[tp]
\centering\includegraphics[width=133mm]{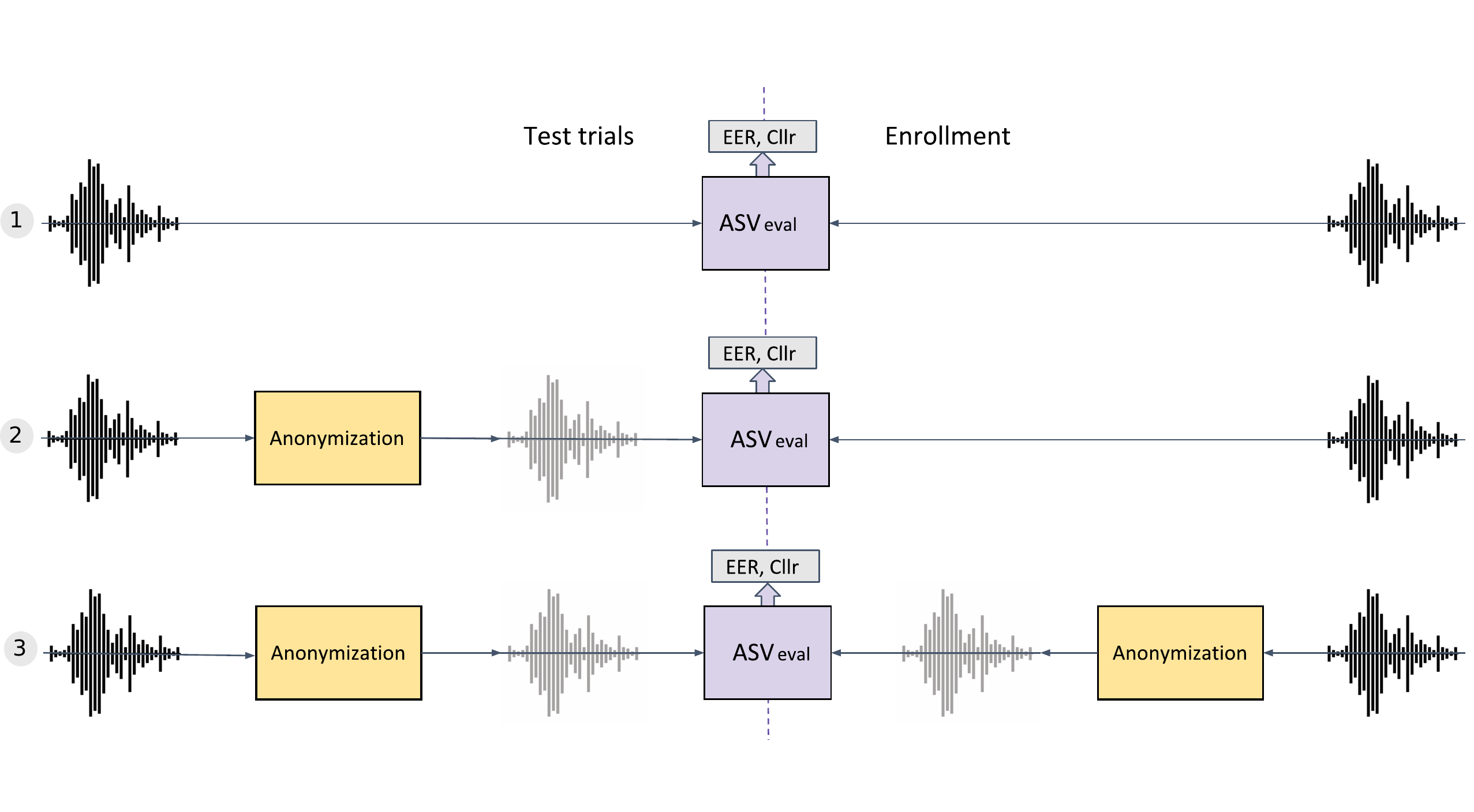}
\caption{ASV evaluation for (1) clean trial and enrollment data; (2) anomymized trial data and clean enrollment data; and  (3) anomymized trial and enrollment data.}
\label{fig:asv-eval}
\end{figure}

\subsubsection{Speaker verifiability metrics}\label{sec:asv-eval} 

The \textrm{$ASV_\text{eval}$} system for speaker verification is based on state-of-the-art x-vector speaker embeddings with a probabilistic linear discriminant analysis (PLDA) backend \cite{snyder2018x} as implemented in Kaldi \cite{povey2011kaldi}. It will be applied as shown in Figure~\ref{fig:asv-eval}:
\begin{enumerate}
    \item Compute PLDA (LLR) scores for (a) clean enrollment data and (b)
anonymized trial data;
    \item  Compute PLDA (LLR) scores for (a) anonymized enrollment data and (b) anonymized trial data;
    \item For steps 1 and 2, estimate the corresponding speaker verifiability metrics: equal error rate (EER) and log-likelihood-ratio cost function ($C_\text{llr}$).
\end{enumerate}
The number of target and impostor trials in the development and evaluation datasets are given in Tables \ref{tab:trials-dev-asv} and \ref{tab:trials-test-asv}, respectively.

\begin{table}[htbp]
  \caption{Number of trials in the \textbf{development} datasets. }\label{tab:trials-dev-asv}
  \renewcommand{\arraystretch}{1.1}
  \centering
  \begin{tabular}{|l|l|r|r|r|}
\Xhline{0.6pt}
 \textbf{Subset} & \textbf{Trials} &  \textbf{Female} & \textbf{Male} & \textbf{Total}  \\ \hline \hline
\multirow{2}{*}{Librispeech: dev-clean}& Target 	& 704	& 644	& \numprint{1348} \\  \cline{2-5}
& Impostor  	&	\numprint{14566} &	\numprint{12796} &	\numprint{27362} \\  \hline \hline
\multirow{4}{*}{VCTK-dev} & Target 	(common part) & \numprint{344}	& \numprint{351}	& \numprint{695} \\  \cline{2-5}
& Target 	(different part) & \numprint{1781}	& \numprint{2015}	& \numprint{3796} \\  \cline{2-5}
& Impostor  	(common part)	&	\numprint{4810} &	\numprint{4911} &	\numprint{9721} \\ \cline{2-5}
& Impostor  	(different part)	&	\numprint{13219} &	\numprint{12985} &	\numprint{26204} \\ \Xhline{0.6pt}
  \end{tabular}
\end{table}

\begin{table}[htbp]
  \caption{Number of trials in the \textbf{evaluation} datasets. }\label{tab:trials-test-asv}
   \renewcommand{\arraystretch}{1.1}
  \centering
  \begin{tabular}{|l|l|r|r|r|}
\Xhline{0.6pt}
 \textbf{Subset} & \textbf{Trials} &  \textbf{Female} & \textbf{Male} & \textbf{Total}  \\ \hline \hline
\multirow{2}{*}{Librispeech: test-clean} & Target 	& 548	& 449	& \numprint{997} \\  \cline{2-5}
& Impostor  	&	\numprint{11196} &	\numprint{9457} &	\numprint{20653} \\  \hline \hline
\multirow{4}{*}{VCTK-test}  & Target 	(common part) & \numprint{346}	& \numprint{354}	& \numprint{700} \\  \cline{2-5}
& Target 	(different part) & \numprint{1944}	& \numprint{1742}	& \numprint{3686} \\  \cline{2-5}
& Impostor  	(common part)	&	\numprint{4838} &	\numprint{4952} &	\numprint{9790} \\ \cline{2-5}
& Impostor  	(different part)	&	\numprint{13056} &	\numprint{13258} &	\numprint{26314} \\ \Xhline{0.6pt}
  \end{tabular}
\end{table}

\subsubsection*{Equal error rate (EER)}\label{sec:eer}

The EER is computed from the PLDA scores as follows. Let $P_\text{fa}(\theta)$ and $P_\text{miss}(\theta)$ denote the false alarm and miss rates at threshold~$\theta$:
\begin{align*}
  P_\text{fa}(\theta)&=\frac{\#\{\text{impostor trials with score} > \theta\}}{\#\{\text{total impostor trials}\}}\\
  P_\text{miss}(\theta)&=\frac{\#\{\text{target trials with score} \leq \theta\}}{\#\{\text{total target trials}\}},
\end{align*}
so that $P_\text{fa}(\theta)$ and $P_\text{miss}(\theta)$ are, respectively, monotonically decreasing and increasing functions of $\theta$. EER corresponds to the threshold $\theta_\text{EER}$ at which the two detection error rates are equal, i.e., $\text{EER}=P_\text{fa}(\theta_\text{EER})=P_\text{miss}(\theta_\text{EER})$.

\subsubsection*{Log-likelihood-ratio cost function ($C_\text{llr}~\text{and}~\text{C}^\text{min}_\text{llr}$)}\label{sec:cllr}

In addition to the EER, an alternative
performance measure, the log-likelihood-ratio cost function ($C_\text{llr}$), will be estimated. It was proposed in~\cite{brummer2006application} as an \textsl{application-independent}\footnote{In this context, \textit{application} is  defined as  a set of prior probabilities and decision costs involved in the inferential process~\cite{brummer2006application}.} evaluation objective and is defined as follows:

\begin{equation*}
C_\text{llr}=\frac{1}{2}\left(\frac{1}{N_\text{tar}}\sum_{i\in \text{targets}}\log_2{\left(1+e^{-\mathrm{LLR}_i}\right)}+\frac{1}{N_\text{imp}}\sum_{j\in \text{impostors}}\log_2{\left(1+e^{\mathrm{LLR}_j}\right)}\right),
\end{equation*}
where $N_\text{tar}$ and $N_\text{imp}$ are respectively the number of target and impostor $LLR$ values in the evaluation set.

$C_\text{llr}$ can be decomposed into a discrimination loss ($C^\text{min}_\text{llr}$)\footnote{How good are two classes separated for any threshold? %$C^\text{min}_\text{llr}$ integrates over all minDCF thresholds $\theta$.
} and a calibration loss ($C_\text{llr}-C^\text{min}_\text{llr}$)\footnote{In the light of domain shifts (e.g., changing speech signal quality between training and evaluation sets), scores loose the capability of correctly encoding the expected class ratio for what their value suggests; scores are not LLRs anymore and thus loose their goodness for ideal decision making. The calibration loss quantifies this gap; scores are considered to be LLRs if $C_\text{llr}\leq1$, and for $C_\text{llr}=1$, a coin toss provides the same information.}~\cite{brummer2006application}.
The $C^\text{min}_\text{llr}$ is estimated by
optimal calibration using monotonic transformation of scores to their empirical LLR values. 
To obtain this monotonic transformation, the pool adjacent violators
(PAV) to LLR algorithm is used~\cite{brummer2006application,ramos2008cross}.

\subsection{Subjective metrics}\label{sec:subj-metr}
\label{sec:perf_subjective}

Several subjective metrics will be used in evaluation: speaker verifiability, speaker linkability, speech intelligibility, and speech naturalness.
They will be evaluated using listening tests that will be carried out by the organizers as described below.

\subsubsection{Subjective speaker verifiability}

To evaluate subjective speaker verifiability, the following two approaches can be used.
The first approach is to measure the speaker similarity between one anonymized trial utterance and one clean enrollment utterance through a large-scale crowdsourced evaluation, following the approach used in a recent study on speech anti-spoofing \cite{wang2019asvspoof}. 
Given one anonymized trial utterance and one randomly selected clean enrollment utterance from the same original speaker, the subjects will be instructed to imagine a scenario in which the anonymized sample is from an incoming telephone call, and the subjects need to judge whether the voice in the call is similar to the clean voice of the claimed speaker. The instruction given to the subjects is as follows:
\begin{quotation}
Imagine you are working for a bank call center. Your task is to compare customer inquiries with voices recorded when the same customer made inquiries in the past. You must determine whether the voices are of the same person or another person who is impersonating the original voices. However, if you choose `spoofing by someone else' more than necessary, there will be many complaints from real customers, which should be avoided. Imagine a situation in which you are working to protect bank accounts and balance convenience.
Now press the `Sample A' and `Sample B' buttons below and listen to the samples. You can listen to them as many times as you like. Use only the audio you hear to determine if the speakers are the same or not. The content of the conversation in English is irrelevant and does not need to be heard. Please judge on the basis of the characteristics of the voice, not the content of the words. 
\end{quotation}
As the instruction describes, the subject will listen to two samples (i.e., Sample A and Sample B) in one evaluation page. Sample B is always a randomly selected enrollment utterance of a given original speaker. Sample A may be an anonymized trial utterance or a clean trial utterance of the same original speaker or a different speaker. The subject will be asked to evaluate the similarity using a scale of 1 to 10, where 1 denotes `different speakers' and 10 denotes `the same speaker' with highest confidence. 
Note that, by using clean speech of the same original speaker or a different speaker as Sample A, we will have anchors in the listening test and can visualize the performance of each participant system through Detection Error Tradeoff (DET) curves.

\subsubsection{Subjective speaker linkability}
The second subjective evaluation metric is based on speaker linkage, i.e., clustering. The listeners will be asked to place a set of anonymized trial utterances
from different speakers in a 1- or 2-dimensional space according to speaker similarity. This will be done using a specially developed graphical interface, where each utterance is represented by a point in the space.  The distance between two points expresses the subjective perception of speaker dissimilarity between the corresponding utterances.

\subsubsection{Subjective speech intelligibility}

Naturalness of the anonymized speech will also be evaluated through a large-scale crowdsourced evaluation. For one subjective evaluation round, the subject will listen to a sample and evaluate its naturalness using a scale from 1 (completely unintelligible) to 10 (completely intelligible and all content is clear). The sample can be an anonymized trial utterance or a clean enrollment utterance from a randomly selected speaker. The results can be visualized through DET curves for comprehensive comparison.

\subsubsection{Subjective speech naturalness}

Similar to the subjective speech intelligibility evaluation, naturalness of the anonymized speech will also be evaluated through a large-scale crowdsourced evaluation. The subject will listen to a sample and evaluate its naturalness using a scale from 1 (completely unnatural) to 10 (completely natural). The sample can be an anonymized trial utterance or a clean enrollment utterance from a randomly selected speaker. The performance of the participants can be visualized through DET curves for comprehensive comparison.

\section{Baseline}\label{sec:baseline}

Two different baseline systems have been developed for the challenge\footnote{Both baseline systems are available online:  \url{https://github.com/Voice-Privacy-Challenge/Voice-Privacy-Challenge-2020}}: (1) anonymization  using x-vectors and neural waveform models, and (2) anonymization using McAdams coefficient.

\subsection{Baseline-1: Anonymization  using x-vectors and neural waveform models}

\begin{figure}[htbp]
\centering\includegraphics[width=130mm]{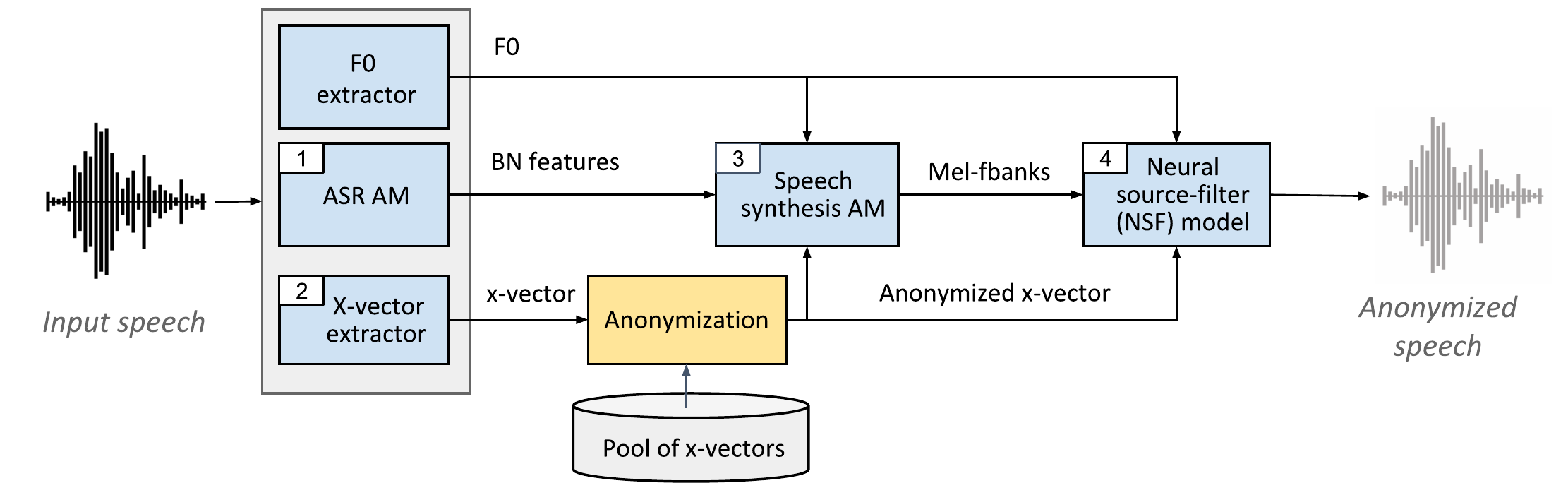}
\caption{Baseline anonymization system.}
\label{fig:baseline}
\end{figure}

Our primary baseline system is based on the speaker anonymization method proposed in~\cite{fang2019speaker} and shown in Figure~\ref{fig:baseline}.
The anonymization is performed in three steps:
\begin{itemize}
\item \textbf{Step 1: Feature extraction:} extract the speaker x-vector~\cite{snyder2018x}, the fundamental frequency (F0) and 
bottleneck (BN) features
from the original audio waveform.
\item \textbf{Step 2: X-vector anonymization:} anonymize the x-vector of the source speaker using an external pool of speakers.

\item \textbf{Step 3: Speech synthesis:} synthesize the speech waveform from the
anonymized x-vector and the original BN and F0 features using an acoustic model and a neural waveform model.
\end{itemize}

\begin{table}[htbp]
  \caption{Baseline anonymization system: models and corpora. The model indexes are the same as in Figure~\ref{fig:baseline}. Superscript numbers represent feature dimensions.}\label{tab:data-baseline-models}
  \label{tab:baseline-models-data} \small
 \renewcommand{\tabcolsep}{0.09cm} 
 \renewcommand{\arraystretch}{1.5}
  \centering
  \begin{tabular}{|l|c|l|l|l|}
\Xhline{0.6pt}
  \textbf{\#} & \textbf{Model} & \textbf{Description} & \textbf{Output features} & \textbf{Training dataset} \\ \hline \hline
  1 & ASR AM & \makecell[l]{ TDNN-F\\  Input:  MFCC$^{40}$ + {\textrm{i-vectors}$^{100}$} \\ 17 TDNN-F hidden layers \\ Output: 6032 triphone  ids \\ LF-MMI and CE criteria \\ }  & \makecell[l]{ BN$^{256}$ \textrm{ features} \\   extracted from \\ the final hidden layer} &  \makecell[l]{Librispeech:\\ train-clean-100 \\ train-other-500} \\ \hline
2 & \makecell{X-vector \\ extractor} & \makecell[l]{TDNN \\ Input: MFCC$^{30}$ \\  7  hidden layers + 1 stats pooling layer \\ Output: 7232 speaker ids \\ CE criterion\\ } & \makecell[l]{ speaker \\ x-vectors$^{512}$} & VoxCeleb: 1, 2 \\ \hline
3 & \makecell{Speech\\ synthesis\\ AM} & \makecell[l]{Autoregressive (AR) network \\ Input: F0$^{1}$ + \textrm{BN}$^{256}$+\textrm{x-vectors}$^{512}$ \\ FF * 2 + BLSTM + AR + LSTM * 2 \\ + highway-postnet \\ MSE criterion }  & \makecell[l]{ Mel-filterbanks$^{80}$} & \makecell[l]{LibriTTS: \\ train-clean-100} \\ \hline
4 & \makecell{NSF \\  model} & \makecell[l]{sinc1-h-NSF in \cite{wang2019neural} \\ Input:  F0$^{1}$ + \textrm{Mel-fbanks}$^{80}$ + \textrm{x-vectors}$^{512}$  \\ STFT criterion} & speech waveform & \makecell[l]{LibriTTS: \\ train-clean-100} \\  \hline \hline
\multicolumn{4}{|c|}{Pool of speaker x-vectors} & \makecell[l]{ 
LibriTTS: \\ train-other-500} \\ \Xhline{0.6pt}
  \end{tabular}
\end{table}

     In order to implement these steps, four different models are required as shown in Figure~\ref{fig:baseline}. Details for training these components are presented in Table~\ref{tab:data-baseline-models}.

In \textsl{Step 1}, to extract BN features, an ASR acoustic model (AM)  is trained (\#1 in Table~\ref{tab:data-train}). We assume that these BN features represent the linguistic content of the speech signal. The ASR AM has a factorized time delay neural network (TDNN-F)  model architecture~\cite{povey2018semi,peddinti2015time} and is trained using the Kaldi toolkit~\cite{povey2011kaldi}.
To encode speaker information, an x-vector extractor with a TDNN model topology (\#2 in Table~\ref{tab:data-train}) is also trained using Kaldi.

In \textsl{Step 2}, for a given source speaker, a new anonymized x-vector is computed by averaging a set of candidate x-vectors from the speaker pool for which the similarity to the x-vector of the source speaker is  in the given range. 
The cosine distance $\cos{(x_1, x_2)}$ or, optionally, probabilistic linear discriminant analysis (PLDA)) distance is used  as a similarity measure between two x-vectors $x_1$ and $x_2$.  
The candidate x-vectors for averaging are chosen in two steps. First, for a given x-vector, $N$ the most farthest candidates from the speaker pool are selected. Second,  a smaller subset of $N^*$  x-vector candidates from this set are chosen randomly\footnote{In the baseline, the following parameter values are used: $N=200$ and $N^*=100$; and PLDA was used as the distance between x-vectors.}.
The x-vectors for the speaker pool are extracted from a disjoint dataset (\textit{LibriTTS-train-other-500}).

In \textsl{Step 3}, 
two modules are used to generate the speech
waveform: a speech synthesis AM that generates Mel-filterbank features
given the F0, the anonymized x-vector, and the BN features, and a neural source-filter (NSF) waveform model~\cite{wang2019neural} 
that produces a speech waveform given the F0, the anonymized x-vector, and the generated Mel-filterbanks.
Both models (\#3 and \#4 in Table~\ref{tab:data-train}) are trained on the same corpus (\textit{LibriTTS-train-clean-100}).

More details about the baseline recipe can be found in the \href{https://github.com/Voice-Privacy-Challenge/Voice-Privacy-Challenge-2020}{provided scripts}.

Results for the ASV objective evaluation are provided in Table~\ref{tab:asv-results} for the development and evaluation datasets. Results for ASR evaluation are presented in Table~\ref{tab:asr-results} in terms of WER.

\begin{table*}[htp]
  \caption{ASV results for \textbf{\textsl{Baseline-1}} for development and test data (\textbf{o} -- original, \textbf{a} -- anonymized speech data for enrollment (\textbf{Enr}) and trial (\textbf{Trl}) parts; \textbf{Gen} denotes speaker gender: \textbf{f} -- female, \textbf{m}~--~male).}\label{tab:asv-results}
  \renewcommand{\tabcolsep}{0.098cm}
  \resizebox{0.95\textwidth}{!}{
  \centering
  \small{
  \begin{tabular}{|c|l|r|r|r||c|c|c||l|r|r|r|}
\Xhline{0.7pt}
 \textbf{\#} & \textbf{Dev. set} & \textbf{EER,\%} &  $\mathbf{C}_{\textbf{llr}}^{\textbf{min}}$  & $\mathbf{C}_{\textbf{llr}}$ & \textbf{Enr} &  \textbf{Trl} & \textbf{Gen}  &  \textbf{Test set}  & \textbf{EER,\%} &  $\mathbf{C}_{\textbf{llr}}^{\textbf{min}}$  & $\mathbf{C}_{\textbf{llr}}$ \\ \hline \hline
1 & libri\_dev & 8.665 & 0.304 & 42.857 & o & o & f & libri\_test & 7.664 & 0.183 & 26.793\\ \hline
2 & libri\_dev & 50.140 & 0.996 & 144.112 & o & a & f & libri\_test & 47.260 & 0.995 & 151.822\\ \hline
3 & libri\_dev & 36.790 & 0.894 & 16.345 & a & a & f & libri\_test & 32.120 & 0.839 & 16.270\\ \hline
\hline 4 & libri\_dev & 1.242 & 0.034 & 14.250 & o & o & m & libri\_test & 1.114 & 0.041 & 15.303\\ \hline
5 & libri\_dev & 57.760 & 0.999 & 168.988 & o & a & m & libri\_test & 52.120 & 0.999 & 166.658\\ \hline
6 & libri\_dev & 34.160 & 0.867 & 24.715 & a & a & m & libri\_test & 36.750 & 0.903 & 33.928\\ \Xhline{0.72pt}
7 & vctk\_dev\_com & 2.616 & 0.088 & 0.868 & o & o & f & vctk\_test\_com & 2.890 & 0.091 & 0.866\\ \hline
8 & vctk\_dev\_com & 49.710 & 0.995 & 172.049 & o & a & f & vctk\_test\_com & 48.270 & 0.994 & 162.531\\ \hline
9 & vctk\_dev\_com & 27.910 & 0.741 & 7.205 & a & a & f & vctk\_test\_com & 31.210 & 0.830 & 9.015\\ \hline
\hline 10 & vctk\_dev\_com & 1.425 & 0.050 & 1.559 & o & o & m & vctk\_test\_com & 1.130 & 0.036 & 1.041\\ \hline
11 & vctk\_dev\_com & 54.990 & 0.999 & 192.924 & o & a & m & vctk\_test\_com & 53.390 & 1.000 & 190.136\\ \hline
12 & vctk\_dev\_com & 33.330 & 0.840 & 23.891 & a & a & m & vctk\_test\_com & 31.070 & 0.835 & 21.680\\ \hline
\hline 13 & vctk\_dev\_dif & 2.864 & 0.100 & 1.134 & o & o & f & vctk\_test\_dif & 4.887 & 0.169 & 1.495\\ \hline
14 & vctk\_dev\_dif & 49.970 & 0.989 & 166.027 & o & a & f & vctk\_test\_dif & 48.050 & 0.998 & 146.929\\ \hline
15 & vctk\_dev\_dif & 26.110 & 0.760 & 8.414 & a & a & f & vctk\_test\_dif & 31.740 & 0.847 & 11.527\\ \hline
\hline 16 & vctk\_dev\_dif & 1.439 & 0.052 & 1.158 & o & o & m & vctk\_test\_dif & 2.067 & 0.072 & 1.817\\ \hline
17 & vctk\_dev\_dif & 53.950 & 1.000 & 167.511 & o & a & m & vctk\_test\_dif & 53.850 & 1.000 & 167.824\\ \hline
18 & vctk\_dev\_dif & 30.920 & 0.839 & 23.797 & a & a & m & vctk\_test\_dif & 30.940 & 0.834 & 23.842\\ \Xhline{0.7pt}
\end{tabular}
}}
\end{table*}

\begin{table}[htp]
  \caption{ASR results for \textbf{\textsl{Baseline-1}}   in terms of WER,\% for development and test data (\textbf{o}-original, \textbf{a}-anonymized speech) for two trigram language models (LMs): \textbf{LM}$_{s}$ - \textrm {small, and }  \textrm{\textbf{LM}}$_{l}$ -  \textrm{large  LM.}}\label{tab:asr-results}
  \centering
  \small
  \begin{tabular}{|c|c|r|r||c||c|r|r|}
\Xhline{0.7pt}
 \multirow{2}{*}{\textbf{\#}}&  \multirow{2}{*}{\textbf{Dev set}} & \multicolumn{2}{c||}{\textbf{WER, \%}}  & \multirow{2}{*}{\textbf{Data}} &  \multirow{2}{*}{\textbf{Test set}} & \multicolumn{2}{c|}{\textbf{WER, \%}} \\ \cline{3-4} \cline{7-8}
 &  & \textbf{\textrm{{LM}$_{s}$}} &\textbf{\textrm{{LM}$_{l}$}} & &   &\textbf{\textrm{{LM}$_{s}$}} &\textbf{\textrm{{LM}$_{l}$}} \\ \hline \hline
1 & libri\_dev & 5.25 & 3.83 & o & libri\_test & 5.55 & 4.15\\ \hline 
2 & libri\_dev & 8.76 & 6.39 & a & libri\_test & 9.15 & 6.73\\  \Xhline{0.72pt}  
3 & vctk\_dev & 14.00 & 10.79 & o & vctk\_test & 16.39 & 12.82\\ \hline 
4 & vctk\_dev & 18.92 & 15.38 & a & vctk\_test & 18.88 & 15.23\\ \Xhline{0.7pt}   
  \end{tabular}
\end{table}

\subsection{Baseline-2: Anonymization using McAdams coefficient}

A secondary, alternative baseline, in contrast to the primary baseline, does not require any  training data and is based upon simple signal processing techniques. It employs the McAdams coefficient~\cite{mcadams1984spectral,patino2020speaker} to achieve anonymisation by shifting the pole positions derived from linear predictive coding (LPC) analysis of speech signals.

\begin{figure}[htp]
    \centering
    \includegraphics[width=122mm]{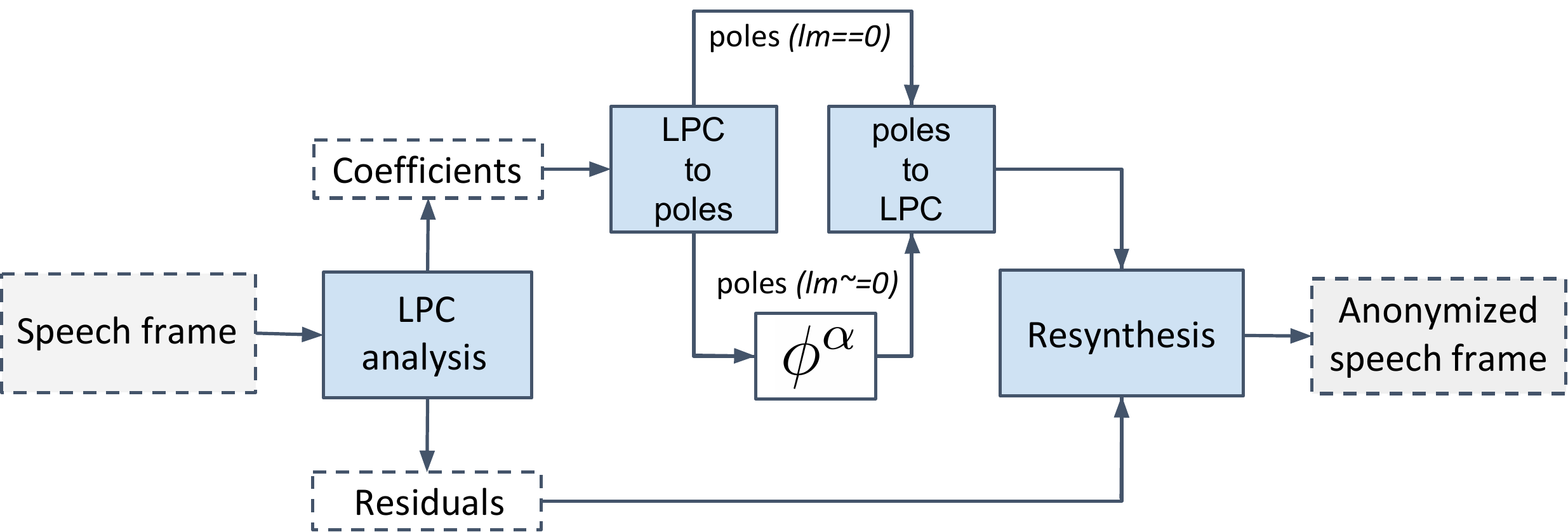}
    \caption{Pipeline of the application of the proposed McAdams coefficient-based approach to anonymisation on a speech frame basis. The angle $\phi$ of poles with a non-zero imaginary part are raised to the power of the McAdams coefficient $\alpha$ to provoke a shift in frequency of its associated formant.}
    \label{fig:lpc_processing}
\end{figure}

\begin{figure}[htp]
    \centering
    \includegraphics[width=\textwidth]{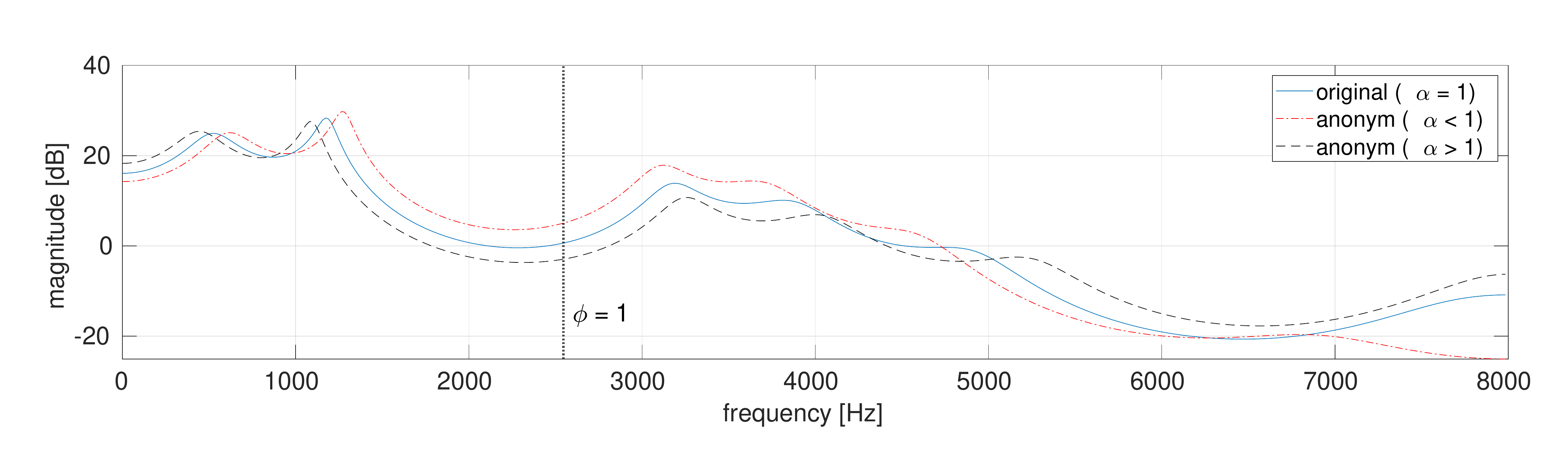}
    \caption{Example of the spectral envelope of a speech frame for both the original formants and the two anonymised versions. }
    \label{fig:mcadams}
\end{figure}

\begin{figure}
    \centering
    \includegraphics[width=0.5\textwidth]{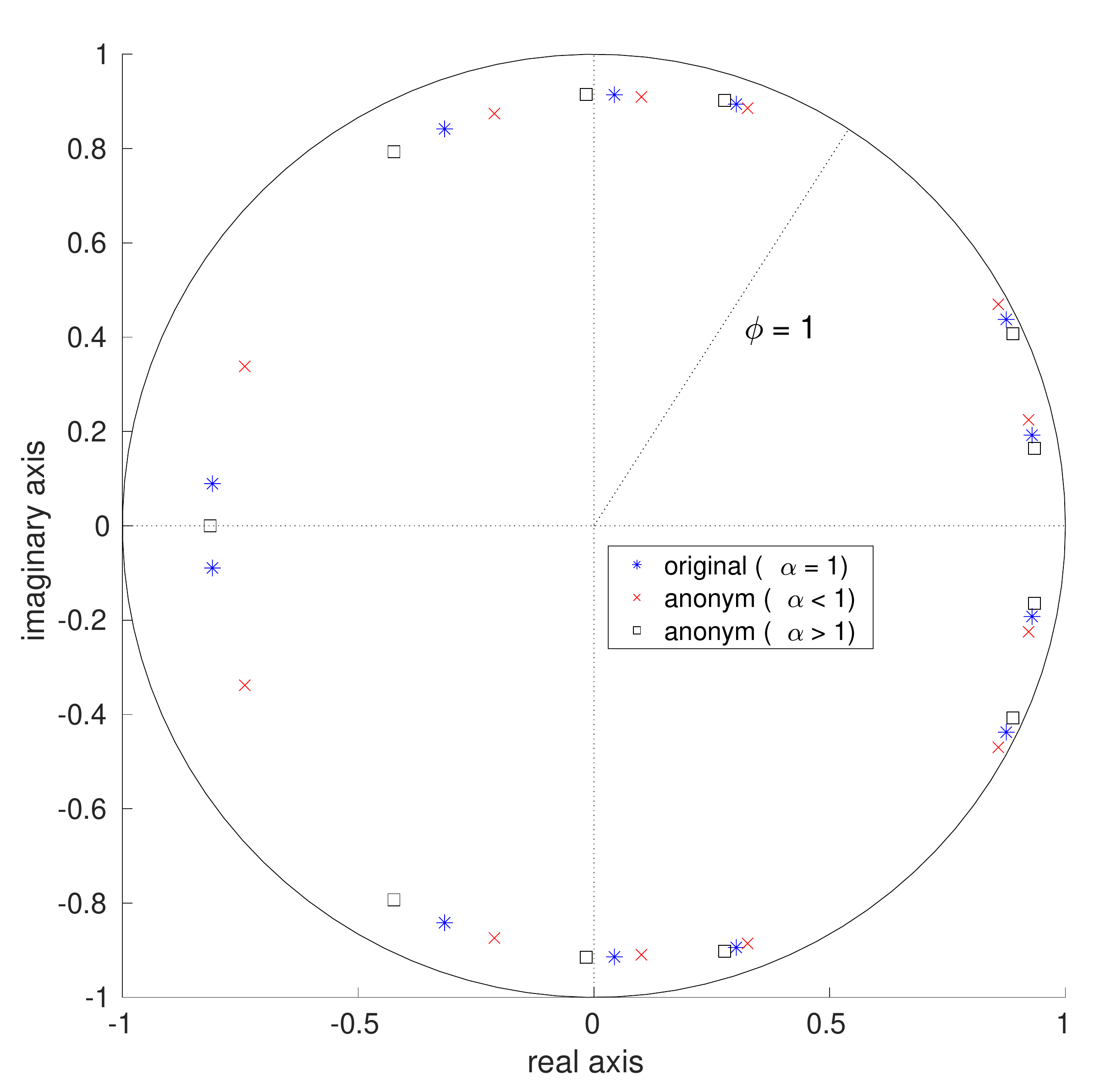}
    \caption{Example of pole-zero map as shown in Figure~\ref{fig:mcadams}}
    \label{fig:pole-zero_plot}
\end{figure}

The process is depicted in Figure~\ref{fig:lpc_processing}.  It starts with the application of frame-by-frame LPC source-filter analysis to derive the LPC coefficients and residuals for a speech signal. Residuals are set aside and retained for later resynthesis. LPC coefficients are converted into a set of pole positions. The McAdams transformation is then applied to the angles of each pole (with respect to the origin in the z-plane), each one of which corresponds to a peak in the spectrum (loosely resembling formant positions).
While real-valued poles are left unmodified, the angles $\phi$ of the poles with a non-zero imaginary part (with values between 0 and $\pi$ radians) are raised to the power of the McAdams coefficient $\alpha$ so that a transformed pole has new angle $\phi^\alpha$. For angles $\phi<1$ (in radians) and $\alpha>1$, $\phi^\alpha$ results in a negative shift in angle, whereas for $\alpha<1$, $\phi^\alpha$ results in a positive shift.  For $\phi>1$, the shift is positive for $\alpha>1$ and negative for $\alpha<1$. The effect of such manipulation upon a set of arbitrary pole positions is illustrated in~Figure~\ref{fig:pole-zero_plot} for values of $\alpha=\{0.9, 1.1\}$. The operation results in the contraction or expansion of the pole positions around $\phi=1$.
The effect upon the corresponding magnitude spectrum is illustrated in Figure~\ref{fig:mcadams}. For a sampling rate of 16kHz, i.e.\ for  data used in the challenge, $\phi=1$ corresponds to approximately 2.5kHz which is the approximate mean formant position~\cite{formant}.
Corresponding complex conjugate poles are similarly and shifted in the opposite direction to their counterparts.  

The new, full set of poles, including original real-valued poles and shifted complex poles are then converted back to LPC coefficients. Finally, LPC coefficients and residuals are used to resynthesise a new speech frame in the time domain. Objective evaluation results for ASV and ASR are illustrated in Tables~\ref{tab:mcadams-ASV-combined} and~Table~\ref{tab:mcadams-ASR-combined} respectively.
This technique is similar in nature to the VoiceMask method~\cite{qian2018hidebehind} that was applied to preserve privacy in speech~\cite{srivastava2019evaluating}. 
While results are inferior to those for the primary baseline, it is stressed that the algorithm has not been optimised in any way.

\begin{table}[tph]
\centering
\small{
%\resizebox{1\textwidth}{!}{
\resizebox{0.95\textwidth}{!}{
\renewcommand{\tabcolsep}{0.098cm}
%\begin{tabular}{|c|l|l|c|c||c|c|c||c|c|c|c|}
%\begin{tabular}{|c|l|r|r|r||c|c|c||l|r|r|r|}
%\Xhline{0.7pt}
\begin{tabular}{|c|l|r|r|r||c|c|c||l|r|r|r|}
\Xhline{0.7pt}
\textbf{\#} & \textbf{Dev. set} & \textbf{EER,\%} &  $\mathbf{C}_{\textbf{llr}}^{\textbf{min}}$  & $\mathbf{C}_{\textbf{llr}}$ & \textbf{Enr} &  \textbf{Trl} & \textbf{Gen}  &  \textbf{Test set}  & \textbf{EER,\%} &  $\mathbf{C}_{\textbf{llr}}^{\textbf{min}}$  & $\mathbf{C}_{\textbf{llr}}$ \\ \hline \hline
1 & libri\_dev & 8.807 & 0.305 & 42.903 & o & o & f & libri\_test & 7.664 & 0.184 & 26.808\\ \hline
2 & libri\_dev & 35.370 & 0.821 & 116.892 & o & a & f & libri\_test & 26.090 & 0.685 & 115.571\\ \hline
3 & libri\_dev & 23.440 & 0.621 & 11.726 & a & a & f & libri\_test & 15.330 & 0.490 & 12.553\\ \hline
\hline 4 & libri\_dev & 1.242 & 0.035 & 14.294 & o & o & m & libri\_test & 1.114 & 0.041 & 15.342\\ \hline
5 & libri\_dev & 17.860 & 0.526 & 105.715 & o & a & m & libri\_test & 17.820 & 0.498 & 106.434\\ \hline
6 & libri\_dev & 10.560 & 0.359 & 11.951 & a & a & m & libri\_test & 8.241 & 0.263 & 15.376\\ \Xhline{0.72pt}
 7 & vctk\_dev\_com & 2.616 & 0.088 & 0.869 & o & o & f & vctk\_test\_com & 2.890 & 0.092 & 0.861\\ \hline
8 & vctk\_dev\_com & 34.300 & 0.877 & 85.902 & o & a & f & vctk\_test\_com & 30.640 & 0.807 & 93.967\\ \hline
9 & vctk\_dev\_com & 11.630 & 0.366 & 43.551 & a & a & f & vctk\_test\_com & 14.450 & 0.465 & 42.734\\ \hline
\hline 10 & vctk\_dev\_com & 1.425 & 0.050 & 1.555 & o & o & m & vctk\_test\_com & 1.130 & 0.036 & 1.042\\ \hline
11 & vctk\_dev\_com & 23.930 & 0.669 & 90.757 & o & a & m & vctk\_test\_com & 24.290 & 0.713 & 99.336\\ \hline
12 & vctk\_dev\_com & 10.540 & 0.316 & 24.986 & a & a & m & vctk\_test\_com & 11.860 & 0.349 & 28.225\\ \hline
\hline 13 & vctk\_dev\_dif & 2.920 & 0.101 & 1.135 & o & o & f & vctk\_test\_dif & 4.938 & 0.169 & 1.492\\ \hline
14 & vctk\_dev\_dif & 35.540 & 0.907 & 90.540 & o & a & f & vctk\_test\_dif & 30.040 & 0.794 & 93.211\\ \hline
15 & vctk\_dev\_dif & 15.830 & 0.503 & 39.811 & a & a & f & vctk\_test\_dif & 16.920 & 0.546 & 41.341\\ \hline
\hline 16 & vctk\_dev\_dif & 1.439 & 0.052 & 1.155 & o & o & m & vctk\_test\_dif & 2.067 & 0.072 & 1.816\\ \hline
17 & vctk\_dev\_dif & 28.240 & 0.741 & 98.419 & o & a & m & vctk\_test\_dif & 28.240 & 0.720 & 101.704\\ \hline
18 & vctk\_dev\_dif & 11.220 & 0.384 & 23.093 & a & a & m & vctk\_test\_dif & 12.230 & 0.397 & 25.064\\ \Xhline{0.7pt}
\end{tabular}
}}
\caption{ASV results for \textbf{\textsl{Baseline-2}} for  development and test data (\textbf{o} -- original, \textbf{a} -- anonymized speech data).}
\label{tab:mcadams-ASV-combined}
\end{table}

\begin{table}[thb]
\centering
\small
%\resizebox{0.7\textwidth}{!}{
\begin{tabular}{|c|c|r|r||c||c|r|r|}
\Xhline{0.7pt}
 \multirow{2}{*}{\textbf{\#}}&  \multirow{2}{*}{\textbf{Dev set}} & \multicolumn{2}{c||}{\textbf{WER, \%}}  & \multirow{2}{*}{\textbf{Data}} &  \multirow{2}{*}{\textbf{Test set}} & \multicolumn{2}{c|}{\textbf{WER, \%}} \\ \cline{3-4} \cline{7-8}
 &  & \textbf{\textrm{{LM}$_{s}$}} &\textbf{\textrm{{LM}$_{l}$}} & &   &\textbf{\textrm{{LM}$_{s}$}} &\textbf{\textrm{{LM}$_{l}$}} \\ \hline \hline
1 & libri\_dev & 5.24 & 3.84 & o & libri\_test & 5.55 & 4.17\\ \hline 
2 & libri\_dev & 12.19 & 8.77 & a & libri\_test & 11.77 & 8.88\\ \Xhline{0.72pt}  
3 & vctk\_dev & 14.00 & 10.78 & o & vctk\_test & 16.38 & 12.80\\ \hline 
4 & vctk\_dev & 30.10 & 25.56 & a & vctk\_test & 33.25 & 28.22\\ \Xhline{0.7pt} 
\end{tabular}
%}
\caption{ASR results for \textbf{\textsl{Baseline-2}} in terms of WER,\% for development and test data (\textbf{o}-original, \textbf{a}-anonymized speech) for two trigram LMs:  \textbf{LM}$_{s}$ - \textrm {small, and }  \textrm{\textbf{LM}}$_{l}$ -  \textrm{large  LM.}}\label{tab:asr-results2}
\label{tab:mcadams-ASR-combined}
\end{table}

\section{Evaluation rules}

\begin{itemize}
    \item Participants are free to develop their own anonymization system, using parts of the baseline or not. They are also free to tune the compromise between speaker verifiability and speech quality/intelligibility     according to their own preference.
    \item  Participants can use only the training datasets and the enrollment datasets specified in Section~\ref{sec:train-data} in order to train their system and tune its hyperparameters. Using additional speech data is not allowed.
    \item Speaker anonymization must be done in a speaker-to-speaker manner. All enrollment utterances from a given speaker must be converted into the same pseudo-speaker, and enrollment utterances from different speakers must be converted into different pseudo-speakers. Also, the pseudo-speaker corresponding to a given speaker in the enrollment set must be different from the pseudo-speaker corresponding to that same speaker in the trial set.
    \item  Modifications of $ASV_\text{eval}$ and $ASR_\text{eval}$ (such as changing decoder parameters) are not allowed.
    \item For every tested system, participants should report all objective metrics (WER, EER, $C_\text{llr}$ and $C^\text{min}_\text{llr}$) on the two development datasets and the two evaluation datasets. Participants should compute these metrics themselves using the provided evaluation scripts. The organizers will be responsible for computing the subjective metrics and the post-evaluation metrics only.
    \item In the system description, participants  should present their results in the same format as used for Tables~\ref{tab:asv-results} and~\ref{tab:asr-results}, or format used in the paper~\cite{tomashenko2020voiceprivacy}.
    Check the following overleaf document for the allowed  table formats to report the results: \url{https://www.overleaf.com/read/cdtgnppxsymj}.
    In order to convert the result file generated  by the evaluation system\footnote{Example: \url{https://github.com/Voice-Privacy-Challenge/Voice-Privacy-Challenge-2020/blob/master/baseline/RESULTS_baseline}}  to {\LaTeX} tables in the required format, participants can use the following script: \url{https://github.com/Voice-Privacy-Challenge/Voice-Privacy-Challenge-2020/blob/master/baseline/local/results_to_latex.py}.

\end{itemize}

\section{Post-evaluation analysis}\label{sec:posteval}

 Following Deadline-2, the organizers will run additional evaluation experiments in order to further characterize the performance of the submitted systems and pave the way for the next Voice Privacy Challenge. To do so, we will ask volunteer participants to share with us the anonymized speech data obtained when running their anonymization system on the training/development datasets and on the evaluation dataset with different settings, and we will compute additional evaluation metrics on these data. The full details of the additional data to be provided by participants and the additional metrics which will be computed will be disclosed in due time.
 
 Roughly speaking, this involves:
\begin{enumerate}
    \item retraining the evaluation systems (\textrm{$ASR_\text{eval}$} and  \textrm{$ASV_\text{eval}$}) on anonymized training data in order to evaluate the suitability of the proposed anonymization technique for ASR and ASV training~\cite{tomashenko2020post};
    \item computing subjective evaluation metrics (as described in Section~\ref{sec:subj-metr}) on selected subsets of evaluation data;
    \item computing additional metrics to assess, e.g., the fulfillment of the goal that pseudo-speakers are sufficiently different from each other and the speaker verifiability performance in the presence of an attacker with additional knowledge.
\end{enumerate}

\section{Registration and submission of results}

\subsection{General mailing list}

All participants and team members are encouraged to subscribe to the general mailing list.  Subscription can be done by sending an email to:

\begin{center} \href{mailto:sympa@lists.voiceprivacychallenge.org?subject=subscribe 2020}{sympa@lists.voiceprivacychallenge.org}
\end{center}

\noindent with \textsl{`subscribe 2020'} as the subject line.  Successful subscriptions are confirmed by return email.  To post messages to the mailing list itself, emails should be addressed to:

\begin{center}
\href{mailto:2020@lists.voiceprivacychallenge.org?subject=voiceprivacychallenge 2020}{2020@lists.voiceprivacychallenge.org}
\end{center}

\subsection{Registration}

Participants/teams are requested to register for the evaluation.  Registration should be performed \textbf{once only} for each participating entity and by sending an email to:

\begin{center}
\href{mailto:organisers@lists.voiceprivacychallenge.org?subject=VoicePrivacy 2020 registration}{organisers@lists.voiceprivacychallenge.org}
\end{center}

\noindent with \textsl{`VoicePrivacy 2020 registration'} as the subject line.  The mail body should include: (i)~the name of the team; (ii)~the name of the contact person; (iii)~their country; (iv)~their status (academic/non-academic).

\subsection{Submission of results}

Each participant may make up to 5 different submissions/algorithms. In case of several system submissions, participants should indicate a primary system among them, and the rest systems should be marked as contrastive. Only primary systems will be considered  in subjective evaluation.
Each single submission should include:
\begin{enumerate}
    \item Metric values for objective assessment (WER, EER, $C_\text{llr}~\text{and}~\text{C}^\text{min}_\text{llr}$) on the evaluation and development datasets according to the common protocol (participants should submit \textsl{result} file obtained by the provided evaluation scripts\footnote{Example of the \textsl{result} file for the baseline system: \url{https://github.com/Voice-Privacy-Challenge/Voice-Privacy-Challenge-2020/blob/master/baseline/RESULTS_baseline}}). 
    \item Corresponding PLDA (LLR) scores  in  Kaldi format (for development and evaluation data) obtained with the provided scripts;
    \item Corresponding anonymized speech  data (wav files, 16kHz,  with the same names as in the original corpus) generated from the evaluation and development  datasets. 
     For evaluation, wav files will be converted to 16-bit Signed Integer PCM format, and this format is recommended for submission.
    These data will be used by the challenge organizers to verify the submitted scores, make post-evaluation analysis with other metrics and to run listening tests for subjective evaluation\footnote{Please note that subjective evaluation will be performed for \textsl{Deadline-2} submissions only, not for \textsl{Deadline-1}}.
\end{enumerate}

All the anonymized speech data should be submitted in the form of a single compressed archive.

Each participant should also submit a single, detailed system description.  All submissions should be made according to the schedule below.  Submissions received after the deadline will be marked as `late' submissions, without exception. 
System descriptions will be will be made  publicly available on the Challenge website.
Further details concerning the submission procedure and URI will be published via the participants mailing list and via the \href{https://www.voiceprivacychallenge.org/}{VoicePrivacy Challenge website}.

\section{Special session at Interspeech 2020}

A special session on the VoicePrivacy 2020 Challenge will be held at  \href{http://www.interspeech2020.org/}{\textbf{Interspeech 2020}} in Shanghai, China. 

All participants are encouraged (but not obliged) to submit their  papers, dedicated to early version of their challenge entry, to the special session of  the conference  by 
\textbf{8th May 2020} (which corresponds to the \textsl{Deadline-1} in the Challenge schedule, see Section~\ref{sec:schedule}).
The review process for papers submitted to the special session is the same as for regular papers. Scientific novelty of the proposed ideas should be clearly stated and evaluated experimentally.
Subjective evaluation results  will be obtained only after \textsl{Deadline-2} and  can be included (optionally) in the camera-ready version of the paper.
Other participants may submit papers to the challenge only.

Accepted Interspeech papers and other Challenge submissions will both be
presented at the special session in the form of posters. These two types of presentations will be
clearly differentiated in the program.

\section{Schedule}\label{sec:schedule}
There will be two submission deadlines (early \textsl{Deadline-1} and late \textsl{Deadline-2}) in the Challenge.
Participants may submit to one or both deadlines. Interspeech paper submission is encouraged but optional. All participants will be invited to present their work at the VoicePrivacy session/event.

\begin{table}[tbh]
  \caption{Important dates}\label{tab:dates}
  \centering
  \renewcommand{\tabcolsep}{-0.12cm} 
  \begin{tabular}{l r }
    \toprule
 Release of training and development data & \textcolor{blue}{8th February 2020} \\ \midrule
 Release of evaluation data  & \textcolor{blue}{15th February 2020} \\  \midrule
 Deadline-1 for participants to submit objective evaluation results & \textcolor{blue}{8th May 2020} \\  \midrule
 Interspeech 2020 paper submission deadline & \textcolor{blue}{8th May 2020} \\  \midrule
 Submission of system descriptions-1 & \textcolor{blue}{15th May 2020} \\  \midrule
 Deadline-2 for participants to submit objective evaluation results and data &	\textcolor{blue}{16th June 2020} \\  \midrule
 Submission of system descriptions-2 &	\textcolor{blue}{23rd June 2020} \\  \midrule
 Submission of additional data  (optional - see Section~\ref{sec:posteval}) & \textcolor{blue}{12d July 2020} \\ \midrule
 Organizers return subjective evaluation results to participants &	\textcolor{blue}{Early/mid August 2020} \\  \midrule
  VoicePrivacy special session/event at Interspeech 2020 & \textcolor{blue}{26th--29th October 2020} \\  \midrule
 Journal special issue submission deadline &	\textcolor{blue}{8th January 2021} \\ \bottomrule
   \end{tabular}
\end{table}

\section{Acknowledgement}
This work was supported in part by the French National Research Agency under projects HARPOCRATES (ANR-19-DATA-0008) and DEEP-PRIVACY (ANR-18-
CE23-0018), by the European Union’s Horizon 2020 Research and Innovation Program under Grant Agreement No. 825081 COMPRISE (\url{https://www.compriseh2020.eu/}), and jointly by the French National Research Agency and the Japan Science and Technology Agency under project VoicePersonae. The authors would like to thank Md Sahidullah and Fuming Fang.

\bibliographystyle{IEEEtran}
\bibliography{main}

\end{document}